\pdfoutput=1
\documentclass[11pt]{article}

\usepackage[]{acl}

\usepackage{times}
\usepackage{latexsym}
\usepackage{colortbl}

\usepackage[T1]{fontenc}
\usepackage[utf8]{inputenc}

\usepackage{microtype}
\usepackage{booktabs}
\usepackage{graphicx}
\graphicspath{ {./graphics/} }
\title{\emph{SimpleStyle}: An Adaptable Style Transfer Approach}

\author{Elron Bandel \qquad Yoav Katz \qquad Noam Slonim \qquad Liat Ein-Dor\\
  IBM Research \\
  \texttt{elron.bandel@ibm.com} \\
  \texttt{\{katz,noams,liate\}@il.ibm.com}}

\begin{document}
\maketitle

\begin{abstract}
Attribute-controlled text rewriting, also known as text style-transfer, has a crucial role in regulating attributes and biases of textual training data and a machine generated text.
In this work we present \emph{SimpleStyle}, a minimalist yet effective approach for style-transfer composed of two simple ingredients:
controlled denoising and output filtering. 
Despite the simplicity of our approach, 
which can be succinctly described with a few lines of code, it is competitive with previous state-of-the-art methods both in automatic and in human evaluation. 
To demonstrate the adaptability and practical value of our system beyond academic data, we apply \emph{SimpleStyle} to transfer a wide range of text attributes appearing in real-world textual data from social networks.
% \liat{from what was said so far, I'm not sure it is clear why masking and sampling are related to our work. I think a connecting sentence is needed here} 
Additionally, we introduce a novel "soft noising" technique that further improves the performance of our system.
We also show that teaching a student model to generate the output of \emph{SimpleStyle} can result in a system that performs style transfer of  equivalent quality with only a single greedy-decoded sample.
Finally, we suggest our method as a remedy for the fundamental incompatible baseline issue that holds progress in the field. We offer our protocol as a simple yet strong baseline for works that wish to make incremental advancements in the field of attribute controlled text rewriting.
\end{abstract}

\section{Introduction}

% The distribution of important attributes of textual data must be carefully controlled at various stages of text processing systems. This includes preprocessing of training data and inputs for models, as well as post processing of model outputs to ensure quality standards. 
\begin{figure}[!t]
    \centering
    \includegraphics[scale=0.45]{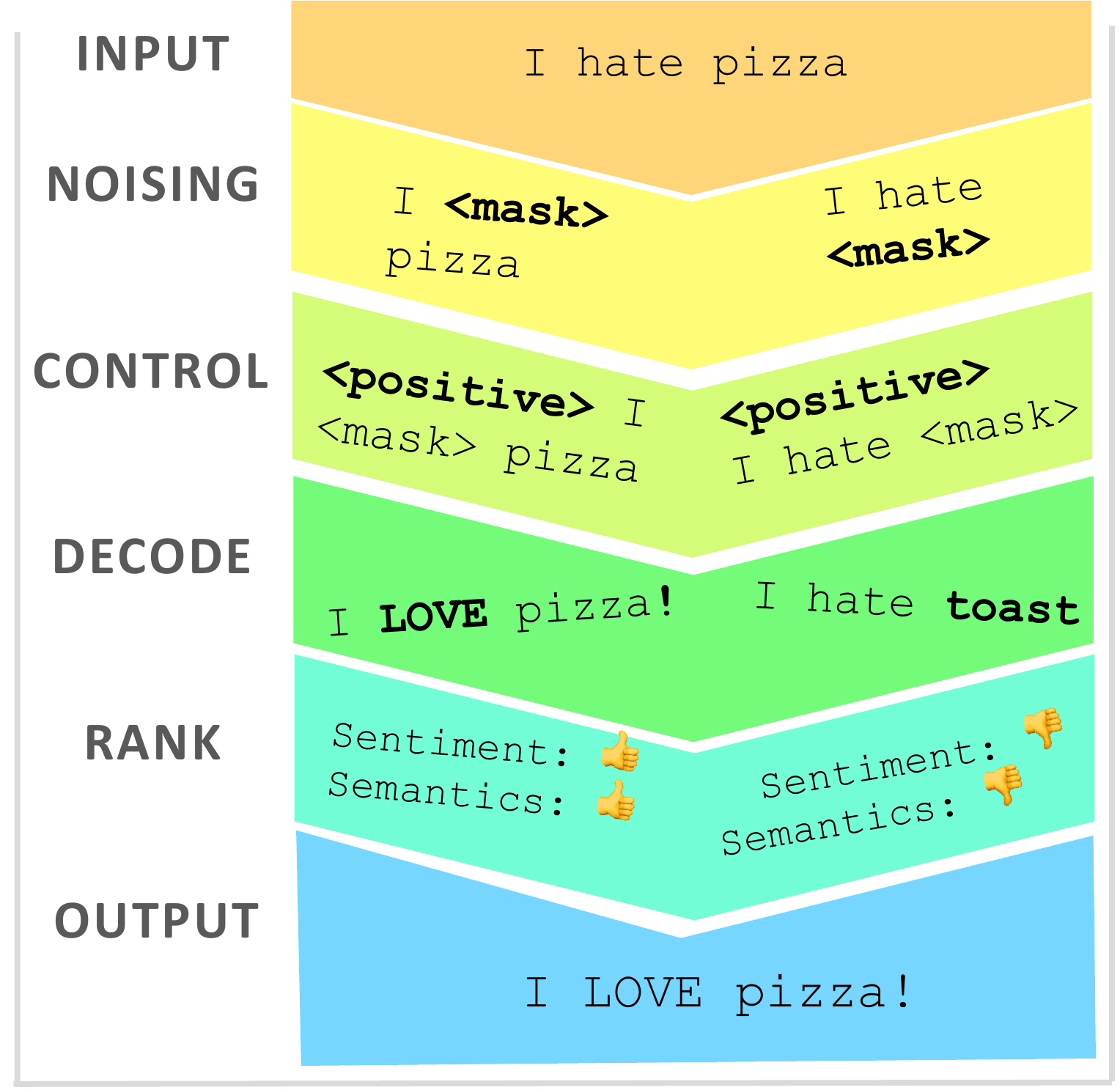}
    \caption{Illustration of \emph{SimpleStyle} sampling process.}
    \label{fig:illustration}
\end{figure}

\begin{figure}[!t]
    \centering
    \includegraphics[scale=0.55]{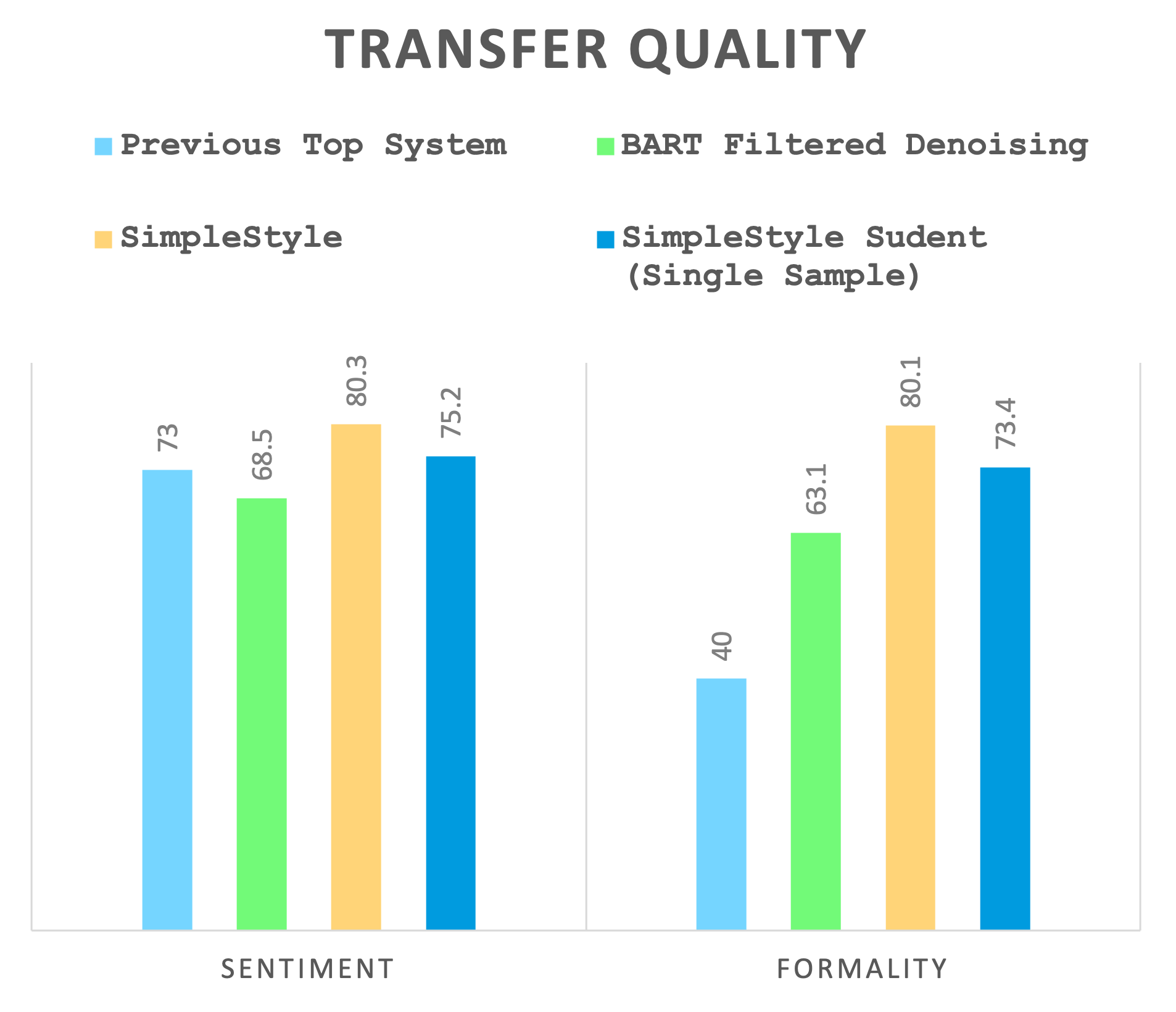}
    \caption{\emph{SimpleStyle} is minimalist yet powerfull in comparison to previous methods, the student of this system can be as strong with only one sample per input.}
    \label{fig:exm}
\end{figure}
 Text attribute transfer is the task of automatically rewriting sentences to possess an attribute such as a specific tense, sentiment and gender, while preserving their attribute-independent content. 
  % Attribute transfer can be a useful tool for a wide range of applications, from content creation to data analysis, and can help users generate high-quality text efficiently and accurately.
  It has recently been established that \textbf{the control of biases and aspects within texts is of paramount importance} throughout various stages of text 
  % processing applications
  classification and generation, \textbf{ranging from the pre-processing of training data to the regulation of output quality} \cite{prabhumoye-etal-2018-style}.
  Due to the shortage of parallel data for attribute transfer,  there is extensive research in developing methods for \textit{unsupervised} attribute transfer \cite{lample2018unsupervised} that do not rely on parallel data.
The two resources that are necessary for developing unsupervised attribute transfer methods, are a \textbf{base model} for text generation and \textbf{automatic metrics} for evaluating the style transfer quality. 
% Automatic evaluation metrics for style transfer must reflect the goal of the system, which is to rewrite the sentence with the desired attribute while minimizing changes to the meaning.

The quality of both models and metrics continuously improve over time, and therefore the  results reported for a method at the time of its inception soon become outdated as more advanced models and metrics are introduced. 
However, it is often impossible to update the baselines established by  old methods using new models and metrics,  
%Training new baselines using old methods is often not possible, 
as these methods often have complex dependencies and may be incompatible with the base model and metrics that are currently being used. 
This creates a major challenge for researchers that wish to evaluate the effectiveness of newly suggested approaches, due to the 
difficulty to disentangle the gain achieved by improved metrics and models from the gains achieved by the approach itself.

To address this issue, it would be desirable to develop an attribute transfer method, which can be easily applied to new base models and automatic evaluation metrics, serving as an adoptable baseline for evaluating the efficacy of a novel approach.

To achieve this goal, we introduce a minimalist approach for style transfer, that 
 can be implemented with just a few lines of code, and relies only on the two necessary aforementioned prerequisites: a base-model and an attribute-classifier that serves as the evaluation metric. 

We show that these two resources are sufficient for achieving a powerful attribute transfer system.

To provide the input base-model with attribute control capabilities with the aid of the metrics, we propose to finetune the model with a controlled version of the masked-sentence-reconstruction task \cite{lewis-etal-2020-bart}. To further improve the model results we suggest to use a classifier-guided sampling of the generated texts. 
The combination of  these two simple components  leads to state-of-the-art results.

% aim to 
% investigate what is the simplest and most stable baseline that can be used for text style transfer.

% Our approach is based on the assumption that every researcher has a base model and evaluation metrics that they intend to use. These are the only ingredients that we assume we have, and we have developed a system that uses only these ingredients, with no additional dependencies.

% To our surprise, despite the simplicity of our approach, our system performs remarkably well.
Overall, our goal is to provide researchers with a reliable and easy-to-use baseline that can help them make more informed and meaningful comparisons of their own style transfer systems, by neutralizing the gains obtained as a result of the updated base model or metrics.
% , the gains from the approach itself from the gains obtained as a result of the updated base model or metrics.
% A comparison to a reference system that is build upon the same model and metrics allows to disentangle the gains of the approach itself from those obtained as a result of the updated base model or metrics. 
We call style-transfer researchers and developers to adopt our minimalist approach, with the hope that it will simplify and assist the development of new style transfer methods.  
 
\section{Related Work}
In the absence of parallel data for supervising the learning process, previous works explored unsupervised solutions for style transfer \cite{DBLP:journals/corr/abs-2010-12742, jin-etal-2022-deep}. 

To establish a simple and reliable baseline for text style transfer, which can be easily retrained by future researchers, we review previous unsupervised methods based on their  \textbf{stability}, \textbf{reproducibility}, \textbf{scalability}, \textbf{reliability} and \textbf{simplicity}.
\paragraph{} \textbf{(1) Stability} -  Early methods that utilized representation bottleneck and adversarial methods to disentangle content and style representations \cite{NIPS2017_2d2c8394, 10.5555/3305381.3305545}, or those that relied on reinforcement learning \cite{ijcai2019p711}, can be difficult to train and suffer from instability \cite{tikhonov-etal-2019-style}. \textbf{(2) Accessibility} - Some methods rely on resources that are not easily accessible, such as specialized lexicons \cite{xu-etal-2016-optimizing, lample2018unsupervised} or output of previous systems \cite{ijcai2019p711}, missing requirements which limits their utility. \textbf{(3) Scalability} - Some methods rely on techniques specific to a certain type of attribute \cite{zhang-et-al-2018-style} or attribute-related sources \cite{lample2018unsupervised}, making them difficult to apply to new attributes. \textbf{(4) Reliability} Some methods are simple but perform well on certain attributes while having poor performance on others \cite{malmi-etal-2020-unsupervised}. \textbf{(5) Simplicity} - Some methods rely on many moving parts with complex relationships that could potentially be simplified.

\paragraph{Denoising Based Methods}
The use of pseudo parallel training data \cite{kajiwara-komachi-2016-building, lee-etal-2019-neural, krishna-etal-2020-reformulating} has shown potential for creating strong baselines that meet many of our requirements. One method for generating such data is to introduce noise to a sentence and train a model to translate the perturbed sentence back to the original, with the expectation that the model will also learn to rewrite unmodified input data\cite{prabhumoye-etal-2018-style}. 

\textbf{Noise Granularity} -
Many text perturbation methods were explored as noising function, starting with the \textsc{identity} function  \cite{xie-etal-2018-noising, Dai2015SemisupervisedSL}, \textsc{character level}  deletion \cite{xie-etal-2018-noising}, insertion \cite{xie-etal-2018-noising} and capitalization \cite{wang-etal-2019-harnessing}).
\textsc{word level} deletions substitutions and masking \cite{ijcai2019p732, malmi-etal-2020-unsupervised, 10.5555/3495724.3496632, lample2018unsupervised, mireshghallah-etal-2022-mix}. \textsc{span level}  deletions \cite{li-etal-2018-delete},  substitutions from lexicon \cite{xu-etal-2016-optimizing} or search  \cite{li-etal-2018-delete}.   \textsc{syntactic level}, perturbing the syntactic structure \cite{zhang-et-al-2018-style}. Lastly, \textsc{sentence level} perturbations, by paraphrasing \cite{xu-etal-2016-optimizing, krishna-etal-2020-reformulating} or back translation \cite{zhang-etal-2020-parallel}. 

We show \textbf{simple span masking is sufficient}. 

\textbf{Attribute-Guided Noising} -
Most of the works that utilize noise try hard to ensure their perturbation techniques alter the attribute, possibly making it a good weak supervision for attribute transfer. Some of the methods do so by relying on external resources ,such as, small set of \textsc{parallel data} for training attribute transferring noising model \cite{xie-etal-2018-noising}, or corpus of similar \textsc{texts without the attribute} that can used for retrieving a text that is similar to the input text  but does not contain the attribute \cite{kajiwara-komachi-2016-building, li-etal-2018-delete, jin-etal-2019-imat, liu2022nonparallel, xu-etal-2016-optimizing, yin2019utilizing}, a \textsc{classifier} that can be used 
for substituting word by word gradually increasing the attribute presence measured by the classifier \cite{10.5555/3495724.3496632, Su_Xu_Qiu_Huang_2018, wu-etal-2019-hierarchical-reinforced, mireshghallah-etal-2022-mix} a \textsc{dictionary} that can be used for replacing text spans with similar text spans without the attribute \cite{zhang-et-al-2018-style, lample2018unsupervised, xu-etal-2016-optimizing,li-etal-2018-delete}. a \textsc{feature attribution system} that is used for identifying spans associated with the attribute that then can be noised or replaced.

In this work we show that there is no need for attribute altering noise, \textbf{random noise is sufficient}. 

\textbf{Model-based Noising} -
One way to alter attributes without relying on external sources is to use a previously trained model, trained by others \cite{ijcai2019p711} or the one currently training \cite{10.5555/3495724.3496632, zhang-et-al-2018-style}. This often enhanced through techniques such as attribute-aware decoding \cite{xu-etal-2016-optimizing} or post decoding filtering \cite{10.5555/3495724.3496632, zhang-et-al-2018-style}. Alternatively, reinforcement learning methods, that can also be seen as using current model for augmenting data, such as those described in \cite{jain-et-al-2019-unsupervised}, can be employed. These methods involve using the outputs of the current model, re-weighting them at the sentence level \cite{gong-etal-2019-reinforcement, ijcai2019p711} or token level \cite{https://doi.org/10.48550/arxiv.2204.07696, zhou-etal-2020-exploring}, and training on the re-weighted data using a weighted maximum likelihood estimation (MLE) method such as REINFORCE, or a gradually filtered MLE \cite{ijcai2019p711}.

This study demonstrates the effectiveness of utilizing a predefined \textbf{non-parametric noising is sufficient}. 

However, we show that retraining the system on its own outputs allows for improved performance in terms of producing good output without the need for many samples and filtering.

\paragraph{Simplicity-Efficiency Trade-offs} In contrast to most previous approaches, our method posits that it is possible to trade efficiency for simplicity. Specifically, during training, we intentionally introduce noise to the input text by randomly masking certain spans in the hope that the critical attributes will occasionally be masked as well. Then, at inference time, we employ an inefficient process of introducing multiple masking variations until we find one that hopefully covers the attribute in question. These strategies enable our system to achieve a high level of simplicity, allowing us to leave out many unnecessary components used by others.

Among the existing approaches, \emph{TextSETTR} \cite{riley-etal-2021-textsettr} bears the closest resemblance to our method, as it employs simple, random text-denoising that is conditioned on the style given by an unsupervised style embedding vector. However, we demonstrate that by conditioning on the style as predicted by a good attribute classifier, we can attain a stronger signal regarding the attribute present in the sentence, thereby enabling more accurate manipulation of the classified attribute.

\section{The Motivation Classifier-Guided Attribute Control}
\label{sec:classifier}

Recently, it has been shown that LLMs can achieve state-of-the-art results in style transfer tasks \citep{reif-etal-2022-recipe}. While these results are impressive, they rely on natural language instructions and cannot be applied to attributes that cannot be described in words. If for example, the attribute is defined to be whether or not an example originates form a specific  corpus or whether it is a challenging example for a given model, then it cannot be easily described in words, but can still be easily defined by a classifier. Therefore, in this study, we focus on Classifier-Guided Attribute Transfer, namely the attribute
% transfer
is defined and the transfer is optimized with respect to a pre-defined attribute classifier.
% as a measure of truth. 
Our total reliance on the attribute classifier neutralizes the influence of the classifier's quality, and enables us 
to focus only on the effectiveness of the attribute transfer system, independent of the classifier's alignment with the true attribute values.
Improvements in the quality of the attribute classifier can be directly translated into better performance of the attribute transfer system in terms of alignment with human judgments. 
% By focusing on attribute transfer as determined by the classifier, we are able to investigate the factors that contribute to the effectiveness of attribute transfer systems independent of the alignment or miss-alignment of this classifier with reality. 

% This allows us to make progress in attribute transfer without being worried about the limitations of the classifier, and to subsequently incorporate improvements in the classifier's performance into our transfer system to more closely align it with the desired attributes being measured.

\section{Method}

\begin{figure}[!t]
    \centering
    \includegraphics[scale=0.4]{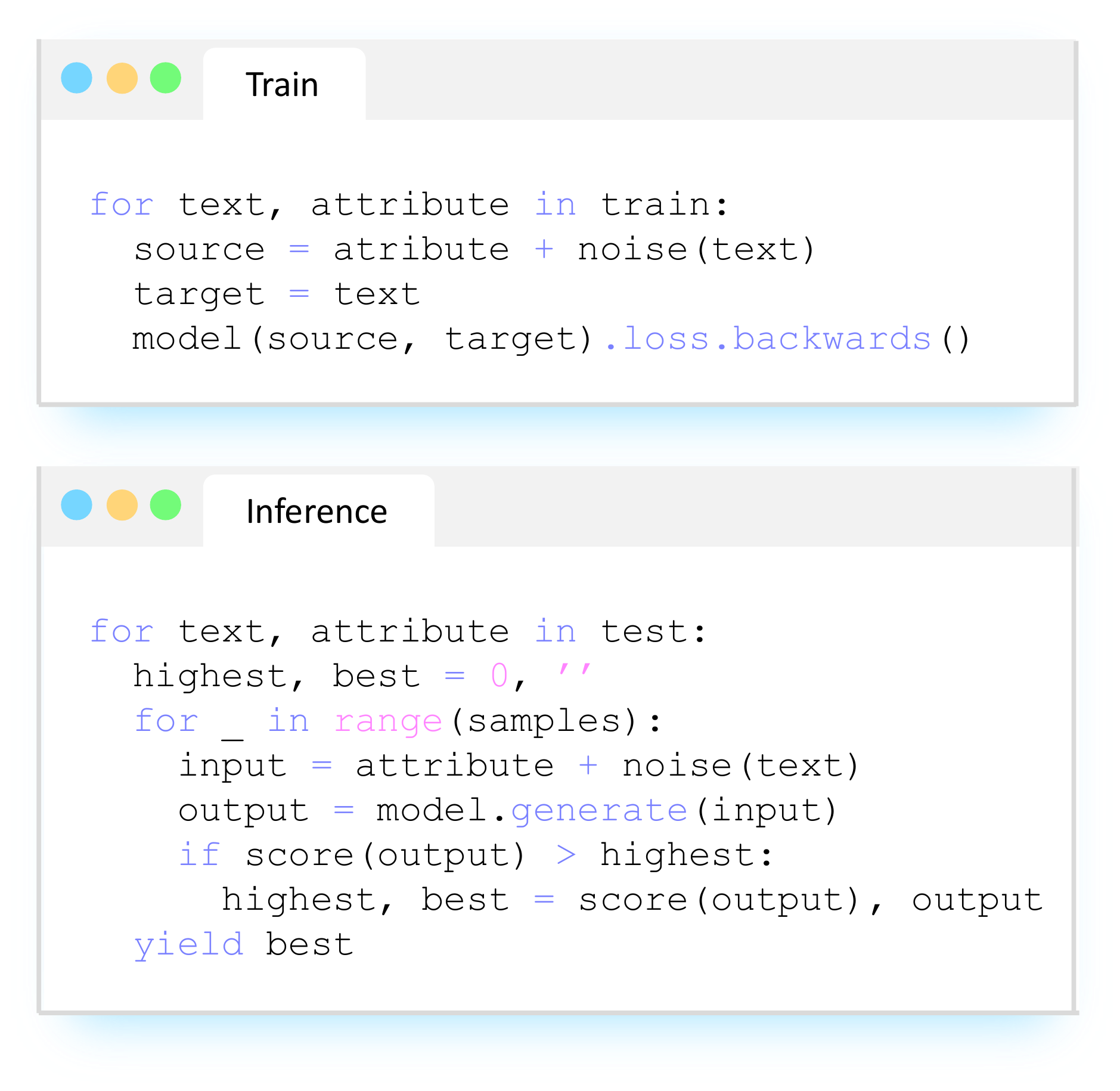}
    \caption{Pseudo code of the \emph{SimpleStyle}  methodology.}
    \label{fig:code}
\end{figure}

Our method, \emph{SimpleStyle}, begins by masking spans of the input text and simply \textbf{finetuning the model to reconstruct the text using a control token} (Figure \ref{fig:code}) that indicates the attribute of the input sentence (as predicted by the attribute classifier).

At inference, several \textbf{masked variations} of the input sentence, concatenated \textbf{with a control token} representing the desired attribute, are fed to the fine-tuned model, aiming \textbf{to generate candidate texts with a transferred attribute}.

% At inference time, we generate several masked versions of the input sentence, and concatenate the token representing the desired attribute to each. 
% % use the attribute token corresponding to the desired attribute, mask the sentence multiple times using hard or soft masks, and generate multiple versions of the unmasked sentence. 
% Each of the resulting versions is fed to the fine-tuned model, to obtain the generated text.
We then use a sentence similarity metric\footnote{We use SimCSE \cite{gao-etal-2021-simcse} because it can be used as an unsupervised semantic similarity metric (with cosine distance), which helps maintain the unsupervised nature of the system making it scalable to new domains} and an attribute classifier to \textbf{select the best candidate}. 
In Section \ref{sec:student} we show that incorporating a student model of \emph{SimpleStyle} can eliminate the need for multiple candidates.

\section{Evaluation}

For evaluating the performance of our system, we follow previous works \cite{xu-etal-2018-unpaired, riley-etal-2021-textsettr} and define the  main evaluation metric, \textbf{\emph{G}}, to be the geometric mean of semantic preservation and attribute level. This metric reflects the goal of the system, which is to \textbf{rewrite the sentence with the desired attribute while minimizing changes to the meaning}. The role of the geometric mean is to penalize cases were there is high success in one aspect and failure in the other. The geometric mean is such cases will be low. 
We turn now to describe the metrics we used in this work to measure each of the quality aspects. 

\paragraph{Semantic} To measure semantic preservation, we use SBERT \cite{reimers-gurevych-2019-sentence} cosine similarity, which has been shown to accurately measure semantic similarity between sentences and correlates well with human judgment \cite{bandel-etal-2022-quality}. 
To measure the semantic preservation we compute the 
cosine similarity between the  sentence embeddings of the input and the human-generated reference, under the assumption that the reference preserves the attribute-independent content of the input.  

\paragraph{Accuracy} For attribute transfer measurement, we use the attribute classifier that defines the attribute, which is also used for filtering at the inference level. This may seem like cheating at first glance, but it is actually an important feature of our setup. As explained in section \ref{sec:classifier}, we view the attribute classifier as the source of truth. As such, we want to make an attribute transfer system that can satisfy the classifier. What if the classifier does not reflect reality? we believe it should be the focus of the classifier developer. progress in classifier defined attribute transfer should be a combination of getting better at transferring the attributes as measured by the classifier and improving the classifier quality in parallel - strengthening the alignment of the system with reality.  While it is not necessarily the case that changing the classifier's prediction will result in an actual attribute transfer in reality, we conduct human evaluations to show that, if you have a good classifier changing the classifier's prediction does indeed result in a satisfying attribute transfer even from a human perspective. We conduct these human evaluations using an A/B testing approach, comparing two systems at a time and asking humans which system is better at (1) transferring the attribute and (2) preserving meaning.
\paragraph{Fluency} To measure how fluent the output sentence is followed by previous works \cite{mireshghallah-etal-2022-mix} we use the perplexity assigned by GPT2 XL as a measure of fluency and grammaticality. 

\paragraph{S-BLEU} To mesure how linguistically similar the output sentence is to the original sentence we use Self-BLEU (S-BLEU) or the BLEU score between the source and the output. This score can give some information about the meaning preservation of sentence but mainly it can measure how much of the original sentence was copied by the model. 

\begin{table*}[t]
\centering
\begin{tabular}{lrrrr>{\columncolor[gray]{0.95}}r}
  \toprule
          \multicolumn{6}{c}{\textbf{Formality}}\\
          \midrule
      \textbf{Method}                 &  \textbf{Fluency}$\downarrow$ &   \textbf{S-BLEU}$\uparrow$&  \textbf{Accuracy}$\uparrow$ &  \textbf{Semantic} $\uparrow$ &  
      \textbf{\emph{G}} $\uparrow$ \\
\midrule
       Copy &   103.93 & 100.00 &   0.07 &      0.79 & 0.14 \\
         Reference &   102.01 &  22.69 &   0.91 &      1.00 & 0.94 \\
         \midrule
         Human1 &   183.08 &  19.01 &   0.87 &      0.72 & 0.77 \\
         Human2 &   227.00 &  19.86 &   0.88 &      0.73 & 0.78 \\
         Human3 &   168.35 &  19.23 &   0.88 &      0.73 & 0.77 \\
         \midrule
             DualRL \cite{ijcai2019p711} &  5417.17 &  16.33 &   0.50 &      0.28 & 0.17 \\
 UnsupervisedMT \cite{prabhumoye-etal-2018-style} &   253.47 &  12.51 &   0.50 &      0.26 & 0.18 \\
\midrule
           \textbf{\emph{SimpleStyle}  (Ours)} &  \underline{65.59} &  50.60 &   \textbf{0.94} &            \textbf{0.69} &       \textbf{0.8} \\
         
          \textbf{\emph{SimpleStyle}  Student (Single Sample)} &  \textbf{53.77} &  46.41 &    \underline{0.84} &          \underline{0.68} &      \underline{0.73} \\
          \toprule
          \multicolumn{6}{c}{\textbf{Sentiment}}\\
          
\midrule
      \textbf{Method}                 &  \textbf{Fluency}$\downarrow$ &   \textbf{S-BLEU}$\uparrow$&  \textbf{Accuracy}$\uparrow$ &  \textbf{Semantic} $\uparrow$ &  
      \textbf{\emph{G}} $\uparrow$ \\
\midrule
    Copy &   266.23 & 100.00 &   0.02 &      0.63 & 0.05 \\
       Reference &   394.11 &  29.52 &   0.81 &      1.00 & 0.84 \\
\midrule
             Back Translation \cite{prabhumoye-etal-2018-style} &   329.47 &   8.07 &   0.95 &      0.29 & 0.50 \\
Delete Generate \cite{sudhakar-etal-2019-transforming} &   677.16 &  42.41 &   0.78 &      0.62 & 0.64 \\
        Constrained Posterior VAE \cite{xu2020variational} &   387.30 &  47.87 &   0.56 &      0.52 & 0.42 \\
            Cross Alignment \cite{NIPS2017_2d2c8394} &   682.73 &  22.10 &   0.74 &      0.39 & 0.47 \\
                  Delete Only \cite{li-etal-2018-delete} &   496.96 &  33.20 &   0.84 &      0.52 & 0.61 \\
              Delete Retrieve \cite{li-etal-2018-delete} &   746.53 &  34.01 &   0.89 &      0.52 & 0.64 \\

                Multi-Decoder \cite{Fu_Tan_Peng_Zhao_Yan_2018} &  1513.17 &  42.55 &   0.48 &      0.41 & 0.30 \\
                   
                Retrieve Only \cite{li-etal-2018-delete} &   290.22 &   5.72 &   0.94 &      0.30 & 0.50 \\
              Style Embedding \cite{Fu_Tan_Peng_Zhao_Yan_2018} &   812.44 &  70.87 &   0.09 &      0.50 & 0.08 \\
                Template Base \cite{li-etal-2018-delete} &  1281.28 &  47.54 &   0.83 &      0.57 & 0.63 \\
                  Cycled RL \cite{xu-etal-2018-unpaired} &  2206.34 &  43.36 &   0.50 &      0.43 & 0.34 \\
                  \midrule
                  MixMatch \cite{mireshghallah-etal-2022-mix} &   411.31 &  52.38 &   0.85 &      0.62 & 0.69 \\
                  DualRL \cite{ijcai2019p711} &   884.17 &  55.02 &   0.88 &      0.66 & 0.72 \\
                UnsupervisedMT \cite{prabhumoye-etal-2018-style} &   717.58 &  42.27 &    \underline{0.94} &      0.60 & 0.73 \\
                \midrule
                  \textbf{\emph{SimpleStyle}  (Ours)} &    \underline{126.86} &  50.51 &   \textbf{0.97} &     \textbf{0.68} & \textbf{0.8} \\
                     \textbf{\emph{SimpleStyle}  Student (Single Sample)} &   \textbf{113.80} &  45.01 &   0.89 &             \underline{0.67} &   \underline{0.75} \\
\bottomrule
\end{tabular}
\caption{The results of the different systems compared to our system. As explained thoroughly in the paper, those results does not imply our system is better, but that it is doing reasonably good. The best score at each category is in bold, and second best is in underline}
\label{tab:results}
\end{table*}

\begin{table*}[t]
\centering
\begin{tabular}{lrrrr>{\columncolor[gray]{0.95}}r}
\toprule
      \textbf{Attribute}                 &  \textbf{Fluency}$\downarrow$ &   \textbf{S-BLEU}$\uparrow$&  \textbf{Accuracy}$\uparrow$ &  \textbf{Semantic} $\uparrow$ &  
      \textbf{G} $\uparrow$ \\
\midrule
  Emotion &   654.47 &  23.41 &   0.87 &      0.64 & 0.73 \\
     Hate &   714.96 &  35.11 &   0.92 &      0.71 & 0.80 \\
    Irony &  1686.44 &  28.18 &   0.90 &      0.70 & 0.78 \\
Offense &  1699.69 &  45.30 &   0.94 &      0.76 & 0.84 \\
Sentiment &  1942.88 &  30.62 &   0.85 &      0.67 & 0.74 \\
\bottomrule
\end{tabular}
\caption{The results of \emph{SimpleStyle}  on transferring \emph{TweetEval} \cite{barbieri-etal-2020-tweeteval} attributes.}
\label{tab:tweet}
\end{table*}

\section{Main Results}

In this section, we present the main results of our study on formality  and sentiment transfer using the most  commonly used benchmark datasets \cite{NIPS2017_2d2c8394, rao-tetreault-2018-dear}. Both datasets consist of non-parallel training data that is divided into groups based on attributes, as well as human-generated parallel test data. As baselines, we employed the top performing popular methods used as baselines in most recent work by \citet{reif-etal-2022-recipe}.

Our automatic evaluation results demonstrate that our system outperforms previous systems in terms of quality in all aspects (see \textbf{Semantic}, \textbf{Accuracy} and \textbf{\emph{G}} in Table \ref{tab:results}). It is worth noting that while we believe our evaluation metrics are effective at evaluating the success of a system , previous systems were trained to optimize different, older metrics. Therefore, it is not necessarily implied that our methodology is superior, but rather that it performs reasonably well. This highlights one of the main points of our paper: without implementing previous methods with our base model and optimizing with the aid of our classifiers and metrics, it is difficult to determine which method is truly the best at learning the transfer as defined by our classifier. Moreover, replicating previous methods is often infeasible due to their complexity. Our method aims to simplify these processes and therefore can serve as a simpler baseline for future researchers. Of course this implies that future adopters of our approach will have to use their updated possibly better classifers and metrics when training our baseline. 

In addition, it is worth mentioning that the attribute-transfer automatic score in the human-annotated test data is not perfect, which might suggest that our classifiers are suboptimal. However, upon reviewing the actual human annotations, we found that our classifier is correct. A random sample exemplifying that point is attached in the appendix. 

Human evaluation further confirmed that our method is comparable to or better than previous methods at transferring attributes as defined by human evaluators.

In the interest of advancing the field, we have made our work transparent by providing a variety of randomly sampled generations from our system in the appendix. Additionally, we have included a subset of comparisons between our method and a previous strong method, without identifying which is which, for the reader to evaluate on their own. In the following page, we will reveal the corresponding outputs for each system.

\begin{table*}
\centering
\resizebox{0.95\linewidth}{!}{%
\def\arraystretch{1.25}\tabcolsep=10pt
\begin{tabular}{p{7.5cm}  p{7.5cm}}

\toprule
\textbf{Input Text} &                                                             \textbf{\emph{SimpleStyle} Generated Output} \\
\midrule
                   totally will come back next time i 'm in town . &                     totally will not return next time i'm in town. \\
                    you can not judge people based on appearance . &       you can tell they care about the people based on appearance. \\
                  i looked around and there were still customers . &   i love the location and the fact that there are still customers. \\
                             the garlic bread was bland and cold . &   the garlic bread was bland and cold and the bread was delicious! \\
      now it is not so thin and taste like any other pizza place . &    now it is not so thin and i love it like any other pizza place. \\
                                 i would n't go back there again . &                            i would definitely go back there again. \\
                          even their steamed veggies are amazing . &                                    even their steamed veggies are. \\
              the charge did include miso soup and a small salad . &      the charge did include miso soup which was very good as well. \\
                               we will spend our money elsewhere . &               we will definitely be spending our money here again! \\
friendly folks , delicious authentic bagels , tasty cream cheese . & friendly folks, delicious authentic bagels, but very poor service. \\

\bottomrule
\end{tabular}}
\caption{Outputs of the \emph{SimpleStyle} system for sentiment transfer.}
\label{tab:generation}
\end{table*}

\section{Ablation Study}
In this section, we conduct an ablation study to examine the impact of various aspects of our approach on the performance of our system.

\begin{figure}[!t]
    \centering
    \includegraphics[scale=0.53]{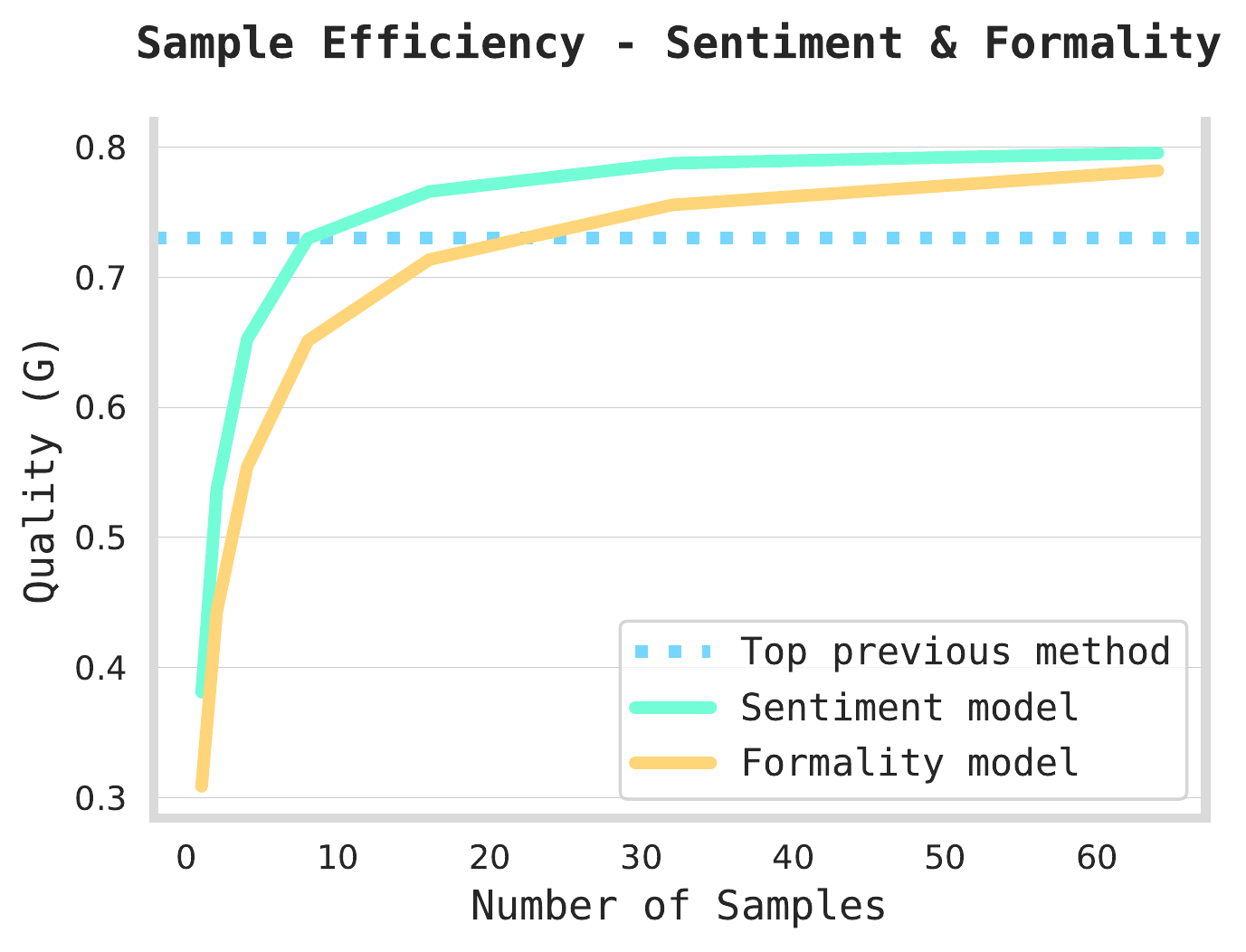}
    \caption{Illustration of the affect of the number of samples followed by filtering on the performance of \emph{SimpleStyle} for both sentiment and formality transfer. }
    \label{fig:sample}
\end{figure}

\paragraph{Number of Samples} Our system is producing at any inference stage few samples from the unmasking \emph{SimpleStyle} model then chosing the best one with the aid of the attribute classifier, firstly how many samples are necessary for achieving high quality outputs? in Figure \ref{fig:sample} we can see that for sentiment tranfer starting at 8 examples our system can achieve SOTA performance, then after 32 examples it reaches plato, while for formality tranfer more smaples might could improver further.

\begin{figure}[!t]
    \centering
    \includegraphics[scale=0.53]{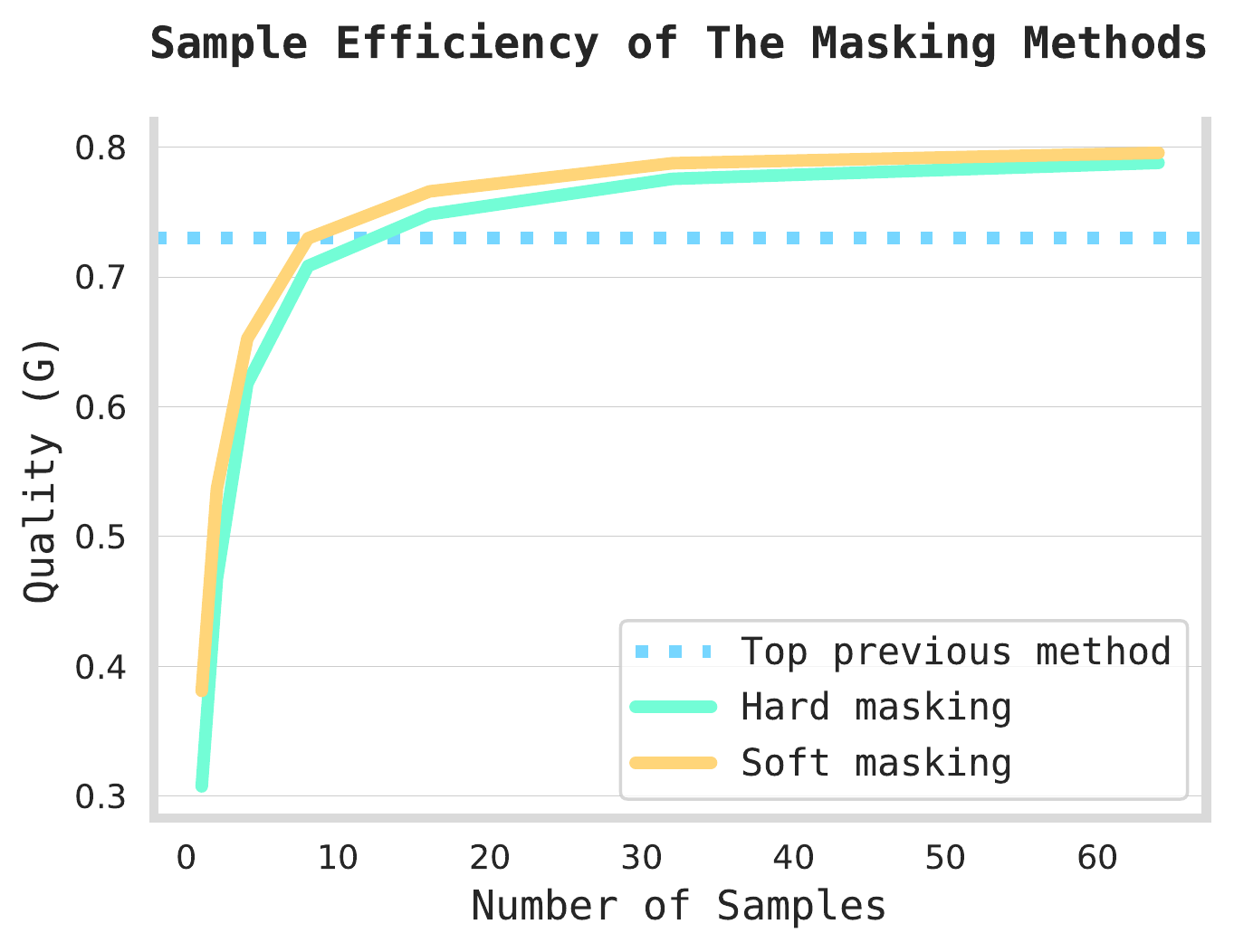}
    \caption{Comparison of the quality of generation of models sampled with soft and hard masking as function of number of samples. It can be seen that soft masking gets higher overall results and is more sample efficient.  }
    \label{fig:hardsoft}
\end{figure}

\paragraph{Soft Masking vs Hard Masking} 
Given a masking budget of covering 0.4 of the output we can either cover words with hard mask, or we can blend their tokens representation vector with the representation vector of the mask token, ie soft masking (inpsired by partial masking for synonim extraction \citet{zhou-etal-2019-bert}). Our results show that for the same masking budgets soft masking is better, and achieve overall better performance (see Figure \ref{fig:hardsoft}). The fact the soft masking does not cover the word completely it allows the model better access to covered tokens and can replace them while preserving some of their meaning. 

\begin{figure}[!t]
    \centering
    \includegraphics[scale=0.53]{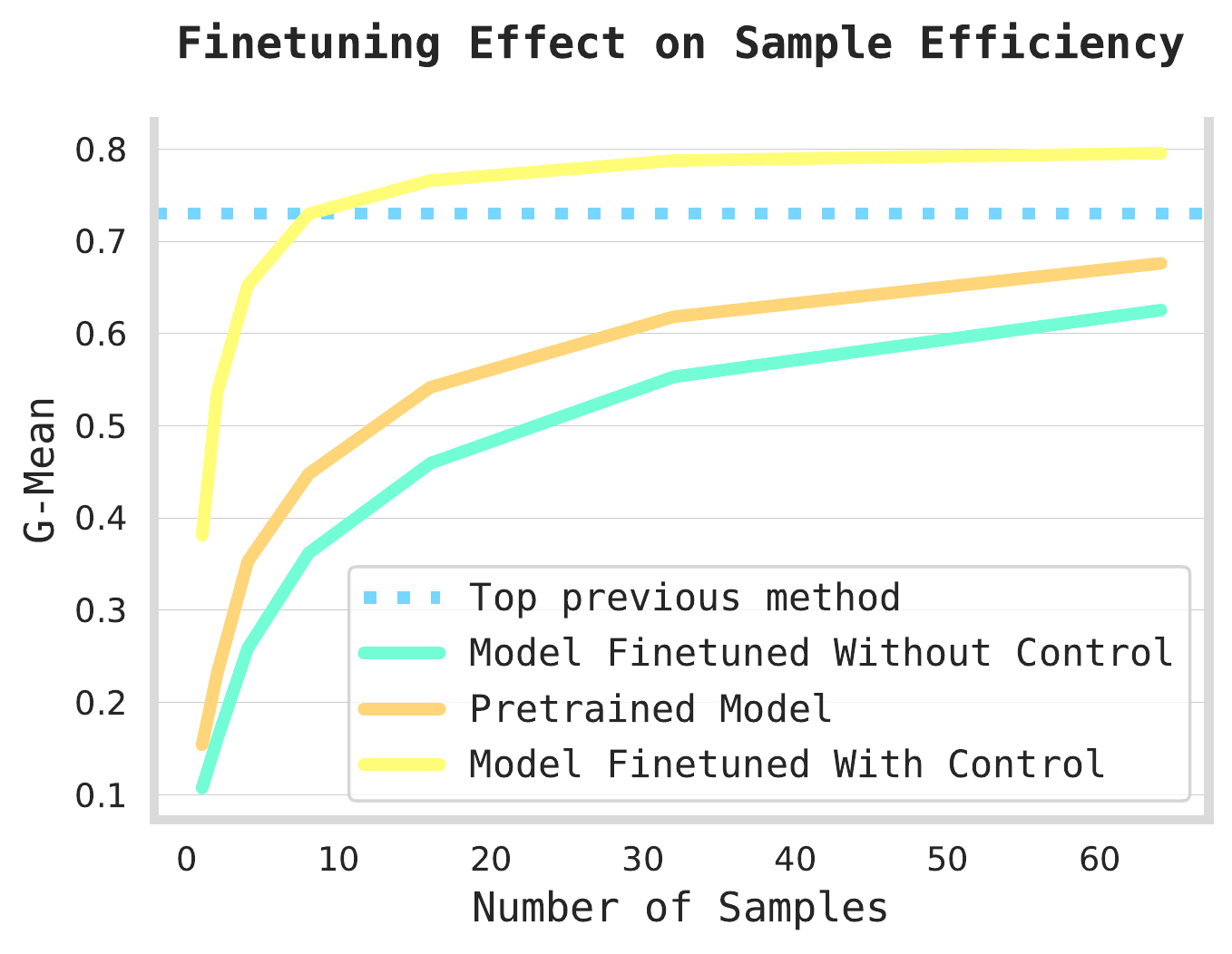}
    \caption{The finetuning with control tokens is necessary for achieving performance higher than top previous system. Any unmasking method without it will not reach that. With as little as 8 samples the performance is already better then previous top methods indicating on better smaple efficiency. }
    \label{fig:pretrained}
\end{figure}

\paragraph{Controlled Finetuned Model VS Vanilla Pretrained} 
how critical is it for the model to be finetuned with the control tokens? could we just sample from vanilla pretrained BART and achieve the same behavior? Recently \citet{mireshghallah-etal-2022-mix} suggested to perform style transfer by filtered sampling from pretrained model, however, our results show, as can be seen in figure \ref{fig:pretrained}, that, in our setup with our metrics, without quality controlled tokens even with extended budget we cannot reach the quality of the baselines, therefore, \textbf{the training scheme we suggest is crucial for high performance end for sample efficiency}. 

\section{Eliminating the Need for Sampling and Filtering with Student Model}
\label{sec:student}
\begin{figure}[!t]
    \centering
    \includegraphics[scale=0.53]{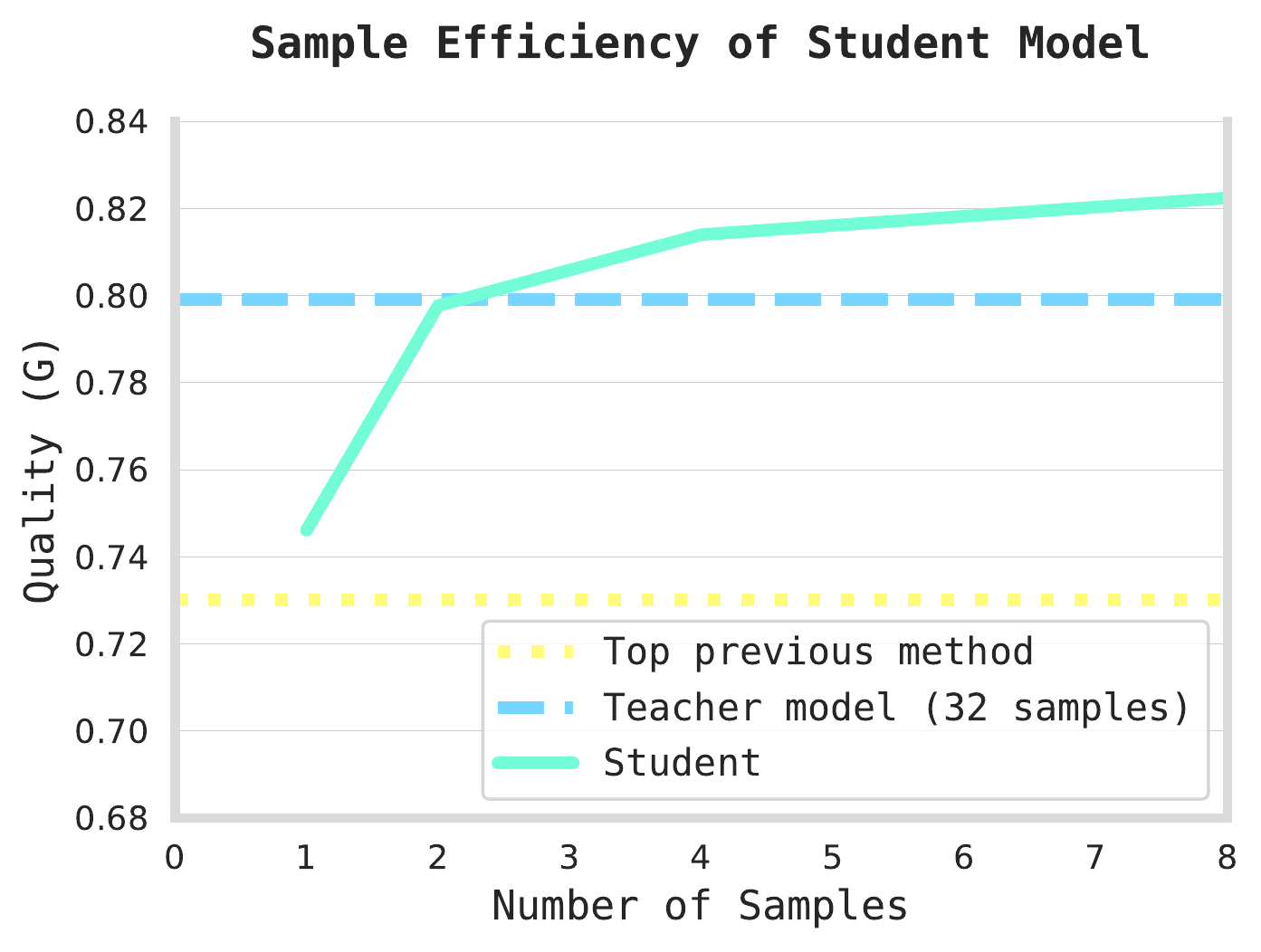}
    \caption{Student model achives with single sample better geometric mean scores of transfer quality metrics (G Score) better then top previous system. With 2 samples and filtering the student model is as good as its teacher, and with 4 samples the student is even better then its teacher.}
    \label{fig:student}
\end{figure}

Sampling the model many times for creating different optional outputs then filtering and choosing the best can be expensive process that is also relying on constant access to the attribute classifier. In order to eliminate the need for this process we took the entire training data and made predictions with our previously proposed system sampled 32 times for every input. Then we trained a student model to produce the predictions of the first system. Our results show that this simple text to text training over the outputs of the original model can create a strong style transfer model. Not only that, this model does not need masking and many samples in order to produce good outputs, with only one sample the model has better performance than the baselines (see Table \ref{tab:results}). more interestingly, with 2 samples and filtering this model is as good as its teacher model,and with 4 samples is even better, suggesting an iterative approach can increase performance even further (see Figure \ref{fig:student}). We also tried to use the teacher as base model for the student but found it is better to initate this model from vanilla bart.  

\begin{figure}[!t]
    \centering
    \includegraphics[scale=0.53]{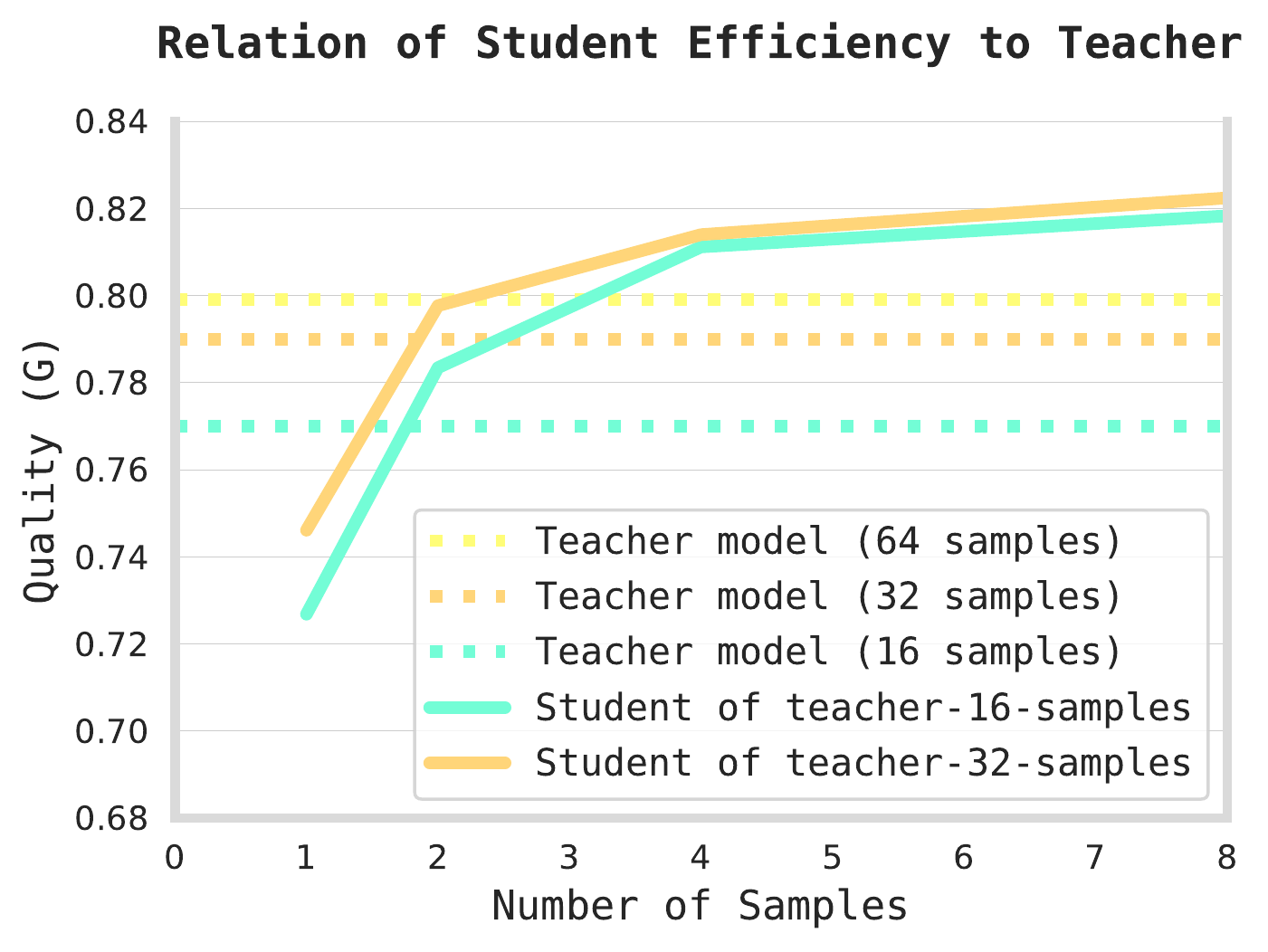}
    \caption{Student model sample efficiency is heavily impacted by the teacher efficiency. Student the learned from better teacher will more likely to be better.}
    \label{fig:student_teacher}
\end{figure}

\section{Scaling Out}
In the last part of this paper we wish to test the effectiveness of our system on real world data taken from social media. This section is crucial for showing the effectiveness of our system when scaled out to other attributes and to harder data distributions that might reflect the relevancy for other researchers.  for this purpose we utilize the \emph{TweetEval} dataset \cite{barbieri-etal-2020-tweeteval} that contain many data for training classifiers of different attributes such as: emotions, irony and hate. \emph{TweetEval} allowed us to experiment with different classifiers trained on data curated carefully. For the classifiers we finetune the large version DeBERTa-v3 model \cite{he2021debertav3} that to the best of our knowledge is the state of the art for classification tasks. Our trained classifiers indeed has the best accuracy based on the online \emph{TweetEval} benchmark, full results are in the appendix.  Then we use Twitter dump of two million tweets augmented with the classifiers predictions for training our system. Our results show that the our approach suceeded in creating attribute transfer system that can transfer the attribute with high accuracy while still maintain the meaning. For example in the case of offensiveness transfer our system achieve attribute transfer accuracy of $0.94$ while achieving $0.76$ SBERT cosine similarity that reflect decent level of meaning preservation. We think that those results should inspire other researchers to use our approach as baseline for their work even if it is complicated attributed and messy real world data.

\section{Conclusions}
In this work, we discuss the importance of having a baseline that is trained with the same base model and optimized with the same metrics as the developed system. The baseline serves as a reference for comparison. We show that with these two basic ingredients - the base model and evaluation metrics - we can develop an extremely simple system that is state-of-the-art. In the ablation study, we demonstrate that each of the components we used was necessary for the success of our system. We also show that when training a student model to replicate the predictions of our system, we can make a style transfer model that is successful with greedy decoding in the first sample. Lastly, we show that our approach can be applied to more attributes with just a corpus of unannotated data and an attribute classifier used for evaluation. Our results suggest that this method will be effective in other scenarios and can serve as the simplest, most accessible, yet powerful baseline for other researchers.

\section{Ethical Considerations}
Attribute controlled text rewriting has the potential to be a powerful tool for editing and regulating the content of text. However, it is important to consider the ethical implications of this technology. While attribute controlled text rewriting can be used to add or remove specific attributes from text, such as hate speech or offensive language, it is crucial to recognize the potential for misuse or abuse. In particular, care should be taken to avoid using attribute controlled text rewriting to amplify or perpetuate hate or harmful messages. At the same time, attribute controlled text rewriting can also be used to reduce hate and promote more positive and inclusive messaging. As researchers and users of attribute controlled text rewriting, it is our responsibility to consider these ethical issues and to act in a way that promotes the responsible and respectful use of this technology.
\clearpage
\bibliography{anthology,custom}
\bibliographystyle{acl_natbib}

\appendix

\section{Appendix}
\label{sec:appendix}

\end{document}